\documentclass{article}
\usepackage{spconf,amsmath,graphicx}
\usepackage{multirow}
\usepackage{enumitem}
\setlist{nosep, leftmargin=14pt}

\usepackage{mwe} % to get dummy images

\usepackage{colortbl}
\usepackage{graphicx}
\usepackage[table,xcdraw]{xcolor}
\usepackage{array}
\usepackage{arydshln}
\definecolor{gray}{gray}{0.7}
\definecolor{lightgray}{gray}{0.7}
\title{VpbSD: Vessel-Pattern-Based Semi-Supervised Distillation for Efficient 3D Microscopic Cerebrovascular Segmentation}
\name{Xi Lin$^{1,2}$ \qquad Shixuan Zhao$^{3,1}$ \qquad Xinxu Wei$^{1,2}$ \qquad Amir Shmuel$^{2,*}$ \qquad Yongjie Li$^{1,}$\sthanks{Corresponding author}}
\address{$^{1}$MOE Key Lab for Neuroinformation, School of Life Science and Technology,\\
University of Electronic Science and Technology of China, Chengdu, China\\
$^{2}$Montreal Neurological Institute, McGill University, Montreal, QC, Canada\\
$^{3}$Department of Neurosurgery, Sichuan Provincial People's Hospital, School of Medicine,\\
University of Electronic Science and Technology of China, Chengdu, China
}
\begin{document}
%\ninept
%
\maketitle
\begin{abstract}
3D microscopic cerebrovascular images are characterized by their high resolution, presenting significant annotation challenges, large data volumes, and intricate variations in detail. Together, these factors make achieving high-quality, efficient whole-brain segmentation particularly demanding. In this paper, we propose a novel \underline{V}essel-\underline{P}attern-\underline{B}ased \underline{S}emi-Supervised \underline{D}istillation pipeline (VpbSD) to address the challenges of 3D microscopic cerebrovascular segmentation. This pipeline initially constructs a vessel-pattern codebook that captures diverse vascular structures from unlabeled data during the teacher model's pretraining phase. In the knowledge distillation stage, the codebook facilitates the transfer of rich knowledge from a heterogeneous teacher model to a student model, while the semi-supervised approach further enhances the student model’s exposure to diverse learning samples. Experimental results on real-world data, including comparisons with state-of-the-art methods and ablation studies, demonstrate that our pipeline and its individual components effectively address the challenges inherent in microscopic cerebrovascular segmentation.
\end{abstract}


\section{Introduction}
\label{sec:intro}
Cerebrovascular segmentation plays a pivotal role in visualizing and extracting intricate details of cerebral blood vessel structures. As a fundamental aspect of neuroscience research, it facilitates comprehensive analysis of brain function and structural changes within cerebral vessels, while also enabling the assessment of vascular abnormalities\cite{poon2023dataset}. These structural changes and abnormalities serve as critical visual indicators for various diseases and degenerative conditions \cite{todorov2020machine, wei2023orientation}. Cerebrovascular malformations, for instance, are lesions stemming from congenital developmental abnormalities, characterized by dynamic structural variations, including localized structural and numerical vessel irregularities\cite{hilton2015neuropathology}. Cerebral small vessel disease refers to brain parenchymal damage induced by diverse structural or functional abnormalities of the brain's small blood vessels\cite{li2018cerebral}. Likewise, cerebral aneurysms alter vascular structures, resulting in functional impairments\cite{jain2007angiogenesis}. 

Given this context, accurately segmenting the entire vascular network, with a specific focus on small vessels, continues to be a principal focus for researchers\cite{chen2023all, wei2023orientation, wei2024retinal}. Small-diameter vessels such as arterioles, capillaries, and venules impose specific requirements on medical imaging modalities to achieve comprehensive segmentation of the cerebral vascular network. High-resolution microscopic cerebrovascular images, characterized by intricate and diverse vascular structures, provide valuable insights for precise analysis in biological research. Therefore, leveraging high-resolution microscopy proves crucial for effectively addressing the challenges associated with cerebrovascular segmentation.

\begin{figure}[tb] % [t]表示图片顶部对齐，可以根据需要调整
    \centering
    \includegraphics[width=0.8\columnwidth]{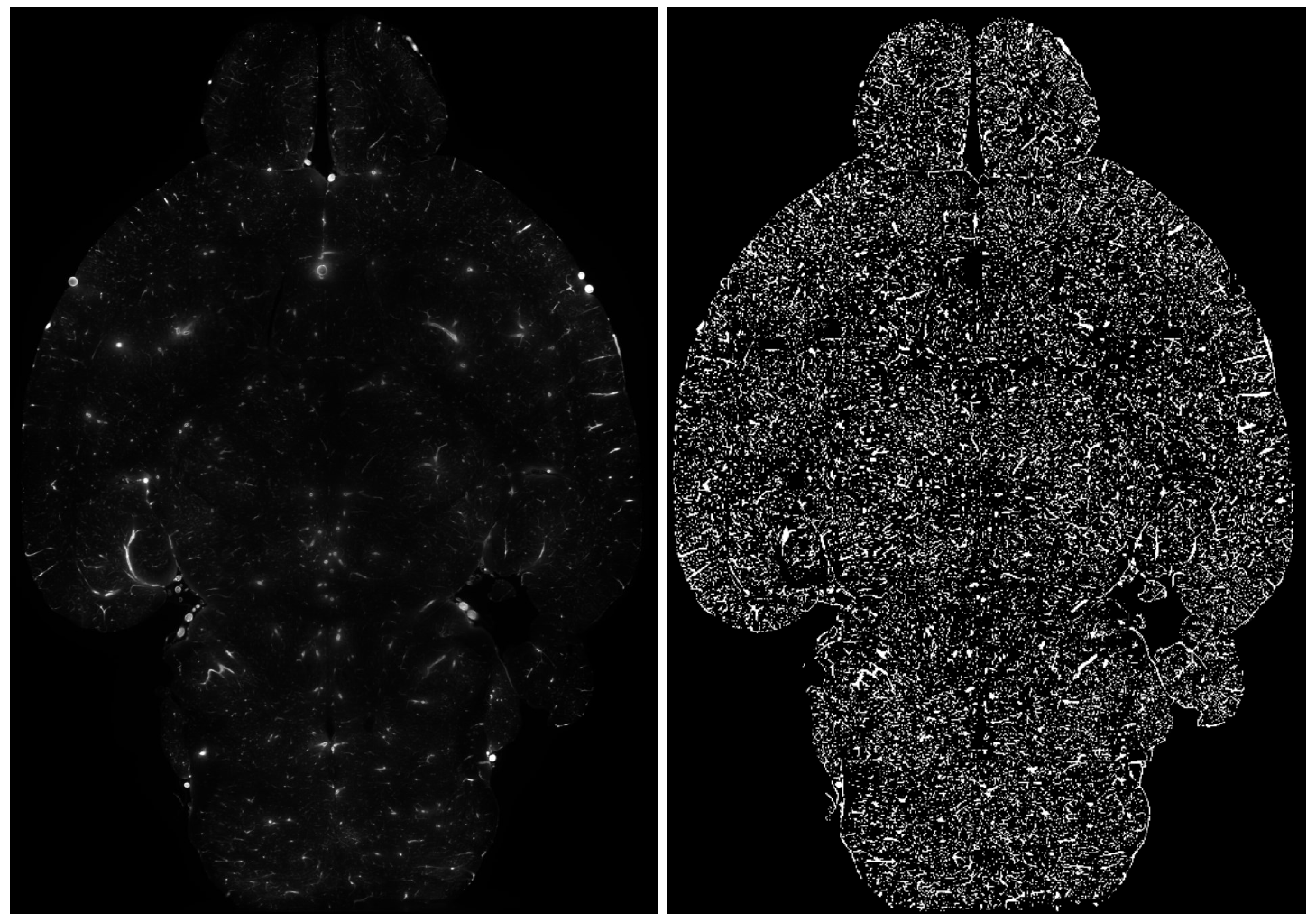} % 将图片宽度设置为文本宽度 
    \caption{A set of 2D segmentation results selected from the final 3D reconstruction of the whole brain segmentation for a CD1 mouse strain.}
    \label{fig:Slice_VIsual}
 %\vspace{-1.5em}
\end{figure}

 However, while some studies have explored cerebrovascular segmentation through imaging modalities such as MRA and CT\cite{chen2022attention,manniesing2006level,banerjee2022topology}, the challenges inherent to microscopic imaging have constrained the development of 3D deep learning approaches for cerebrovascular analysis, leading to relatively few contributions in this area. The high resolution of microscopic images results in each cerebral vascular slice containing millions of pixels. When expanded to the task of 3D whole-brain vascular segmentation, the data for a single 3D microscopic vascular sample can easily reach hundreds of gigabytes. Consequently, the size of an entire dataset may soar to terabyte levels, presenting significant challenges for storage, annotation and processing\cite{todorov2020machine}.  The difficulty of obtaining detailed annotations, along with variability in the shape and size of cerebral vessels and multiple branching levels at their termini, introduces substantial obstacles to the implementation of deep learning methods for cerebrovascular segmentation\cite{farajzadeh2021multiscale,chen2023all}. Additionally, the considerable file sizes associated with high-resolution microscopic images not only render the segmentation of an entire brain using large models exceedingly time-consuming but also necessitate substantial computational resources, leading to significant memory consumption\cite{tetteh2020deepvesselnet,todorov2020machine}.

\iffalse
Despite these challenges, considerable opportunities for further research exist, underscoring the importance of continued exploration. Given the limitations of finely annotating microscopic brain vascular images, leveraging information from unlabeled data presents an effective approach to enhancing model performance. Self-supervised learning techniques—such as cloze tests, jigsaw puzzles, and reconstruction—utilize large volumes of unlabeled data to train models on valuable features without extensive annotations \cite{tang2022self,he2022masked,noroozi2016unsupervised}. Similarly, semi-supervised learning trains models with a small amount of labeled data alongside a larger pool of unlabeled data, strengthening model robustness \cite{tarvainen2017mean,yu2019uncertainty,vu2019advent}. To address this, optimizing the trade-off between segmentation accuracy and processing speed is essential. Knowledge distillation strategies offer an intuitive solution, as they can reduce model size and computational complexity \cite{gou2021knowledge}.
\fi

In response to these challenges, we developed a novel approach VpbSD for 3D microscopic cerebrovascular segmentation. This pipeline consists of two main stages: the pretraining stage for the teacher model and the subsequent stage of knowledge transfer and distillation. Initially, VpbSD employs a self-supervised pretrained teacher model to capture diverse vascular structures from extensive unlabeled data. The self-supervised learning technique leverages large volumes of unlabeled data to train models on valuable features without the need for extensive annotations\cite{tang2022self,he2022masked,noroozi2016unsupervised}. Then, by integrating semi-supervised learning with a novel codebook-based feature knowledge distillation, VpbSD effectively enriches the information acquisition of the student model while ensuring fast and precise segmentation. This approach not only reduces model size and computational complexity but also enhances overall performance. Our work makes the following key contributions:
 
\begin{itemize}

\item First, based on the characteristics of 3D microscopic cerebrovascular data, we design a novel \underline{V}essel-\underline{P}attern-\underline{B}ased \underline{S}emi-Supervised \underline{D}istillation pipeline (VpbSD) for 3D cerebrovascular segmentation, which reduces computational complexity while effectively leveraging unlabeled microscopic data to enhance performance. 

\item Second, we develop a vessel-pattern based knowledge distillation approach that constructs a vessel-pattern codebook during the teacher model's pretraining phase. This codebook enables the teacher model to capture intricate vascular structures from unlabeled data and facilitates efficient knowledge transfer to the student model, thereby improving segmentation accuracy through the sharing of learned vascular diversity and structural context.

\item Third, we evaluate our approach on real datasets, comparing it with other state-of-the-art methods and conducting dilation study, demonstrating the effectiveness of our strategy.

\end{itemize}

The reminder of this paper is organized as follows: Section \ref{sec:rlw} reviews strategies and methods relevant to this work, including microscopic cerebrovascular segmentation, knowledge distillation, self/semi-supervised learning, and  vector quantization. Section \ref{sec:method} provides a detailed description of the proposed cerebrovascular segmentation approach. Section \ref{sec:exp} presents experimental validations. Finally, Section \ref{sec:clu} concludes the paper.

\section{related works}
\label{sec:rlw}

\subsection{ Microscopic Cerebrovascular Segmentation}

There have been some works at segmenting cerebral vessels in microscopic images. \cite{todorov2020machine} developed a pipeline based on transfer learning for cerebrovascular segmentation in light-sheet microscopic images of mice. By pretraining on synthetic datasets and fine-tuning on a small amount of real, annotated data, this approach achieved good segmentation results using a small network. %MiniVess is a high-quality two-photon fluorescence microscopic dataset of mouse cerebrovasculature, published in \cite{poon2023dataset}, with detailed annotations. 
Based on features from microscopic images, deep learning methods\cite{menten2023skeletonization,tetteh2020deepvesselnet,shit2021cldice}, attempt to enhance segmentation results by utilizing vascular morphological characteristics, thereby reducing the burden on equipment.

However, these efforts have not fully leveraged the abundant unlabeled microscopic cerebrovascular data to explore its distributional characteristics. Consequently, there remains a limited amount of work specifically addressing 3D microscopic cerebrovascular segmentation, indicating a significant opportunity for further research in this field.

\subsection{Knowledge Distillation}

The knowledge distillation paradigm, introduced in 2015 \cite{hinton2015distilling}, addresses challenges such as reducing model training costs, safeguarding private data, and optimizing limited computational resources. It involves transferring various forms of "knowledge" from a teacher model to a student model and has evolved into numerous effective variants \cite{gou2021knowledge}.

Knowledge distillation can be categorized into two main types based on the nature of the transferred knowledge: output layer distillation and intermediate layer distillation. Output layer knowledge consists of logits or pseudo labels predicted by the teacher model, encapsulating implicit reasoning about the data \cite{hinton2015distilling, muller2019does}. In contrast, intermediate layer distillation involves transferring intermediate outputs from the teacher model to the student model, providing richer semantic information. A key challenge in this process is aligning the outputs of heterogeneous teacher and student models. Notable contributions in this area include works such as \cite{heo2019knowledge, ahn2019variational, romero2014fitnets, liu2022cross}.

In the task of 3D microscopic cerebrovascular segmentation, the high resolution of images results in substantial storage requirements and computational demands, potentially reaching terabyte levels \cite{todorov2020machine}. Knowledge distillation mitigates these issues by enabling the transfer of knowledge from a complex teacher model to a smaller, faster student model, thereby enhancing computational efficiency while improving representation quality and accuracy during inference \cite{gou2021knowledge}.

\subsection{Self/Semi-supervised Learning}
Unsupervised learning (USL) and semi-supervised learning (SSL) effectively reduce the burden of extensive dataset annotation.
USL facilitates feature discovery by guiding models to learn the data distribution through various tasks. In computer vision, popular self-supervised approaches include image reconstruction \cite{michelucci2022introduction}, masked autoencoders \cite{he2022masked}, jigsaw puzzles \cite{chen2021jigsaw, noroozi2016unsupervised}, and image rotation \cite{taleb20203d, gidaris2018unsupervised}, as well as combinations of these methods\cite{tang2022self}. Pixel-level reconstruction tasks compel models to capture fine-grained features and structural details, which enhances their ability to generalize to downstream applications.

SSL leverages a small set of labeled data and a large volume of unlabeled data to optimize learning\cite{jiao2023learning}. It can be combined effectively with knowledge distillation, as in the Mean Teacher model\cite{tarvainen2017mean}. SSL has also been advanced by integrating approaches like uncertainty estimation\cite{yu2019uncertainty}, entropy minimization \cite{vu2019advent}, and cross pseudo supervision \cite{chen2021semi}. This method is particularly useful in medical image segmentation, where labeled data is limited and costly to obtain\cite{you2022simcvd,you2022momentum,basak2023pseudo}.

In the context of 3D microscopic cerebrovascular segmentation,where labeling data is highly time-intensive, both SSL and USL prove effective in capturing intricate vascular features with minimal reliance on labeled examples. These methods enhance model performance and adaptability, making them well-suited for tackling complex segmentation tasks.

\subsection{Vector Quantization}
Vector quantization was first integrated with unsupervised deep learning in Vector Quantized Variational Autoencoders (VQ-VAE) \cite{van2017neural}. This approach projects the latent continuous encoding space into a discrete vector space defined by an explicitly constructed codebook. The decoder then generates the final output based on the discretized embeddings from the codebook.
A common challenge in VQ-VAE is codebook collapse, where only a limited subset of codebook embeddings is utilized during training, leading to inefficient representation learning and diminished model performance \cite{yu2021vector}. Solutions such as Beit-v2 \cite{peng2022beit} have effectively addressed this issue by employing a dimensionality-reduced codebook. Reg-VQ \cite{zhang2023regularized} promotes broader utilization of codebook vectors through regularized quantization, while VQ-Wav2Vec \cite{baevski2019vq} employs Gumbel-Softmax to make the discretized sampling process continuous, thereby enhancing codebook utilization.

In our 3D microscopic cerebrovascular segmentation task, constructing a codebook serves as a mechanism for acquiring diverse vessel patterns through unsupervised learning, effectively bridging the gap for knowledge distillation.

\section{Methods}
\label{sec:method}
\begin{figure*}[t] % [t]表示图片顶部对齐，可以根据需要调整
    \centering
    \includegraphics[width=\textwidth]{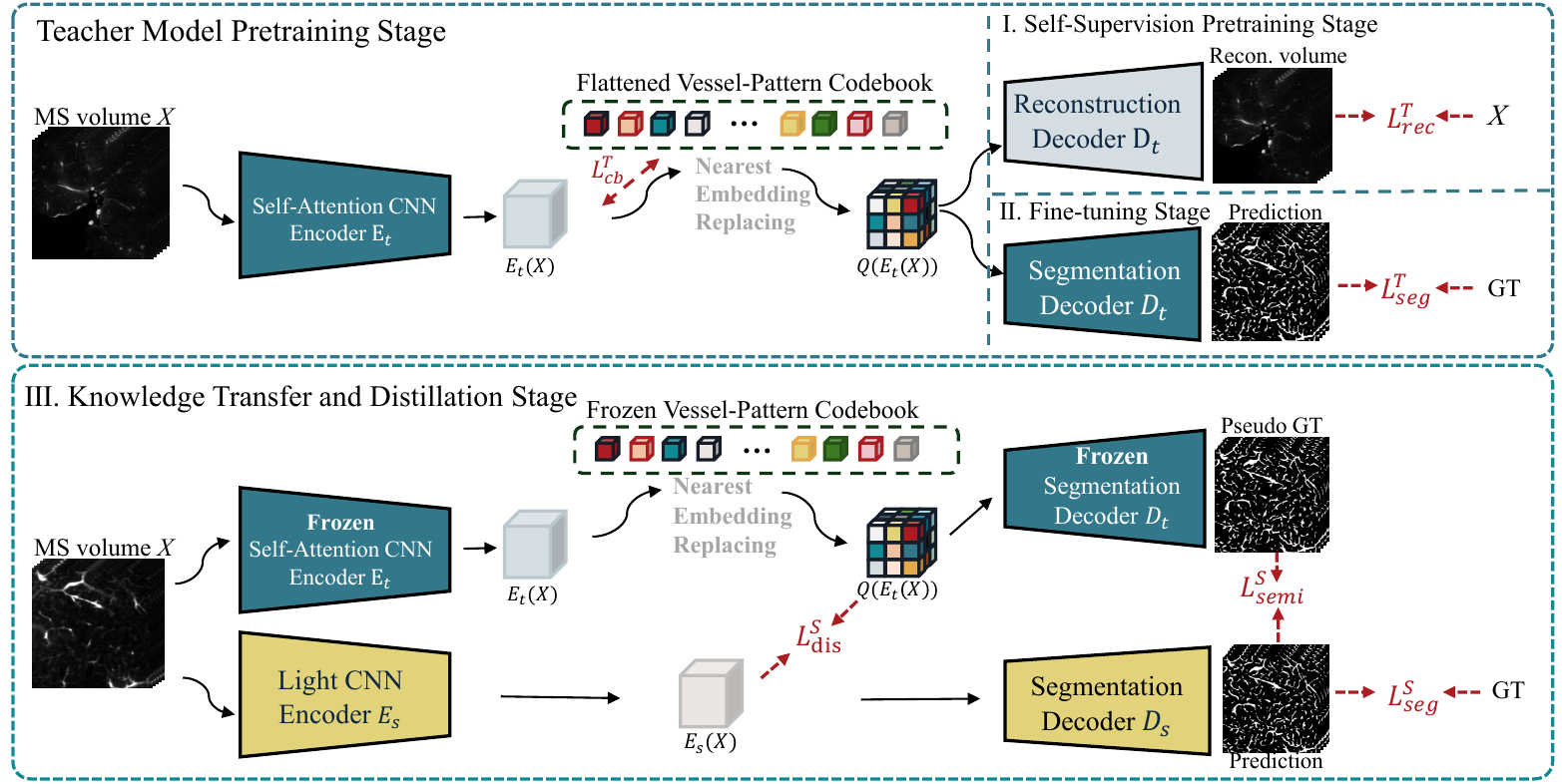} % 将图片宽度设置为文本宽度
    \caption{The training workflow of VpbSD. The black arrows indicate the direction of forward propagation, while the red arrows represent regulatory signals. During the self-supervised pretraining phase of the teacher model, the unlabeled microscopic (MS) volume$X$ is input into the model, which trains by computing the codebook loss $L_{cb}^T$ and reconstruction loss $L_{rec}^T$. In the fine-tuning phase,labeled volumes are employed to calculate the fully supervised segmentation loss $L_{seg}^T$. In the knowledge distillation phase, the full supervision segmentation loss $L_{seg}^S$ is computed for labeled data, while a semi-supervised loss $L_{semi}^S$  is calculated for unlabeled data using pseudo ground truth from the teacher model and predictions from the student model. Additionally, a distillation loss $L_{dis}^S$ is computed between the quantized encoded results from the teacher model and the encoded results from the student model.}
    \label{fig:pipeline}
% \vspace{-1.5em}
\end{figure*}

High-resolution microscopic images present considerable challenges for 3D brain vascular segmentation due to their substantial storage requirements, increased annotation complexity, and prolonged computation times, all of which render the process more resource-intensive. To mitigate these challenges, we introduce a novel codebook-based semi-supervised knowledge distillation framework that addresses these issues concurrently. This section first elucidates the foundational concepts of VQ-VAE, followed by a comprehensive overview of the entire pipeline. Subsequently, we provide an in-depth explanation of each stage and its key components. The training workflow of the proposed method is illustrated in Fig.\ref{fig:pipeline}.

\subsection{Preliminary Knowledge}
For VQ-VAE, let $X$ represent the input image, $E(*)$ denote the encoder, $D(*)$ the decoder, and $Q(*)$ the quantizer. The inference process of VQ-VAE can be described as follows:
\begin{equation}
    VQ-VAE(X)=D(Q(E(X)))
\end{equation}
Furthermore, let $\mathcal{E}=\{e_i \mid i=1,2,…,m\}$ denote the original output of the encoder, and let $\mathcal{V}_{n \times d}=\{v_j \mid j=1,2,…,n\}$ represent the embedding space of the codebook, where $n$ indicates the size of the codebook and $d$ represents the dimensionality of the embeddings.  The inference process of the quantizer, which updates $e$ to the quantized vector $e^{'} $ using the vectors from the codebook, can be described as follows:
\begin{equation}
    Q(\mathcal{E}) = \{e^{'} _i =v_j\mid \underset{j}{argmin} \parallel e_i - v_j\parallel,i=1,2,…,m\}
\end{equation}
In VQ-VAE, the quality of the generated image is significantly influenced by the quality of the codebook. Consequently, the loss function comprises two components: reconstruction loss and quantization loss. Since discrete selection prevents gradients from flowing, a straight-through estimator \cite{bengio2013estimating} is employed to address this issue. Thus, the quantization loss can be expressed as:
\begin{equation}
\mathcal{L}_{\text{quant}} = \left\| \text{sg}[\mathcal{E}] - v_i \right\|_2^2 + \lambda \left\| \mathcal{E} - \text{sg}[v_i] \right\|_2^2
\end{equation}
where $sg(*)$ is the stop-gradient operation.
\subsection{Overall Framework}
VpbSD comprises two primary stages: the pretraining of the teacher model and the subsequent knowledge transfer and distillation phase.

In the pretraining phase, we utilize a robust self-attention CNN teacher model that is pretrained on unlabeled data to effectively capture the distribution characteristics of the dataset. Concurrently, a vascular pattern codebook is developed. This choice of a teacher model, which integrates both self-attention and CNN mechanisms, enables the learned codebook to encompass not only local vascular structural patterns but also contextual information regarding the vasculature.
During the knowledge transfer and distillation phase, we freeze all parameters of the teacher model and apply semi-supervised learning to train a lightweight CNN student model using a combination of unlabeled and annotated data. The codebook acts as a conduit for knowledge transfer from the teacher model, thereby enhancing the student model's ability to integrate features. This results in a lightweight yet high-precision student model capable of performing rapid segmentation.

\subsection{The Teacher Model Pretraining Stage}
The pretraining phase of the teacher model consists of two sub-stages: the self-supervision pretraining stage and the fine-tuning stage. In the first stage, the teacher model learns diverse vascular pattern features from a substantial set of unlabeled microscopic volumes. The second stage involves fine-tuning the model using a limited amount of labeled data to yield high-quality segmentation results. Ultimately, the well-trained teacher model provides rich knowledge for the student model.
\subsubsection{ The Self-Supervision Pretraining Stage}
In 3D microscopic cerebrovascular datasets, even a single slice may contain millions of pixels due to high resolution, resulting in datasets that often require terabytes of storage \cite{todorov2020machine}. Manually annotating such extensive data can take years, rendering the task impractical \cite{todorov2020machine}. Therefore, minimizing reliance on annotated data while fully leveraging the distribution characteristics learned from unlabeled images becomes paramount.

Our approach employs a teacher model that utilizes self-supervised learning to reconstruct vascular images while concurrently developing a vessel-patterns codebook that captures essential vascular structures. This discrete codebook enables effective representation of intricate vascular features, which are subsequently transferred to the student model through knowledge distillation, thereby enhancing both segmentation accuracy and efficiency.

The teacher model adopts an encoder-decoder architecture. To ensure that the learned vascular patterns preserve local morphological features and contextual spatial relationships, we incorporate a self-attention CNN encoder. This design captures fine-grained details and broader contextual information, ensuring comprehensive feature representation throughout the segmentation process. Notably, both the encoder and decoder are flexible and interchangeable components, allowing for adaptability across various model configurations.

The loss function comprises both reconstruction loss and codebook loss. We utilize the mean squared error (MSE) loss for reconstruction. The overall loss function for pretraining can thus be expressed as:
\begin{align}
L_{pre}^T &= L_{rec}^T + \alpha \cdot L_{cb}^T \\
&= \text{MSE}(X, X') + \alpha \cdot  L_{cb}^T \\
&= \text{mean}_{i} \left( \left| x_{i} - y_{i} \right| \right) + \alpha \cdot  L_{cb}^T
\vspace{-1.5em}
\end{align}
Upon completion of self-supervised learning, the teacher model is deemed to have captured the intrinsic characteristics of the data, enabling it to effectively analyze and extract complex vascular morphological features.
\subsubsection{ The Fine-Tuning Stage}
The fine-tuning phase follows a standard fully supervised segmentation training paradigm. Given that the teacher model has been pretrained on self-supervised tasks, we assume it has become familiar with the dataset's distribution characteristics and is capable of effectively analyzing vascular patterns.  Consequently, by utilizing a small set of labeled cerebrovascular data and performing fine-tuning over several epochs, we can obtain a high-quality segmentation teacher model. %During this phase, the codebook remains frozen and does not undergo further training, while only the decoder is updated.

The loss function used in this part is the commonly applied Dice loss for vascular segmentation, specifically: 

\begin{equation}
L_{seg}^T  = \frac{2\left | Y_{tea}\cap Y_{gt} \right |  }{\left | Y_{tea} \right | +\left | Y_{gt} \right | } 
\end{equation}
Here, $Y_{tea}$ represents the model's output, and $Y_{gt}$ denotes the vascular labels.

\subsection{Vector Quantize based Knowledge Distillation Strategy}

\begin{figure}[tb] % [t]表示图片顶部对齐，可以根据需要调整
    \centering
    \includegraphics[width=\columnwidth]{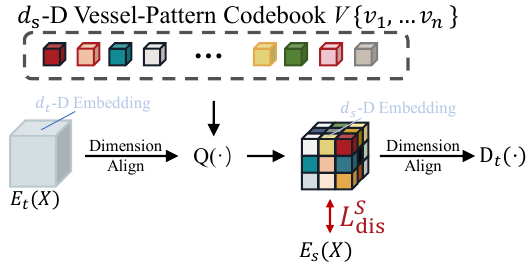} % 将图片宽度设置为文本宽度
    \caption{ An illustration of the vessel-pattern-based knowledge distillation module.}
    \label{fig:vpbKD_Module}
\end{figure}

In this work, we introduce a novel vector quantization-based heterogeneous model knowledge distillation strategy to effectively address the complexities inherent in 3D microscopic cerebrovascular data, which poses significant challenges for comprehensive annotation. Despite the difficulty of annotation, unlabeled data inherently contain diverse vascular structure distribution information, which can be leveraged to enhance the model's generalization ability, thereby benefiting the construction of comprehensive brain vascular segmentation results.

As previously discussed, self-supervised learning can extract valuable insights from this unlabeled data. However, a key challenge lies in effectively transferring the rich vascular pattern features acquired by the teacher model during self-supervised training to the student model. Given the student model's limited learning capacity, maximizing the quality of the learned features is essential for conveying meaningful information. Furthermore, the inherent heterogeneity between the teacher and student models—characterized by the teacher's more complex structure and significantly greater number of parameters—creates disparities in their intermediate layer outputs. Organizing this knowledge to facilitate effective learning for the student model presents another substantial challenge.

To address these challenges, we propose a codebook-based heterogeneous model knowledge distillation strategy.  By utilizing vector quantization, we replace the original encoding vector with the closest matching vector from the codebook, transforming the encoding space from continuous to discrete \cite{baevski2019vq,van2017neural}.  The codebook, learned during the self-supervised learning phase of the teacher model, serves as an explicit storage structure for vascular pattern information. It captures the diverse distribution characteristics of the entire rat cerebrovascular network, enhancing the model's generalization ability. The codebook integrates complex and varied vascular structural features, providing refined information. This transformation allows the intermediate layer knowledge of the teacher model to shift from continuous and infinitely variable outputs to finite, discretized information, thus making it more robust against noise and reducing the learning difficulty for the student model. Additionally, using discrete semantic vectors instead of original pixel-level vectors facilitates a more global and structured learning of the dataset features.

Due to the teacher model's encoder possessing superior encoding capabilities and a more intricate structure, it generates output vectors of higher dimensionality. To facilitate knowledge integration in a manner that is more accessible for the student model, the dimensions of our codebook do not directly align with the output dimensions of the teacher model's encoder. Instead, we employ a dimension alignment method to map the outputs of the teacher model's encoder to dimensions compatible with those of the student model's encoder, resulting in a lower-dimensional codebook.  Let the embedding dimension of the teacher model’s encoder output be $E_t(X) = Z\{z_1,...z_j\}$  with a dimensionality of $d_t$, and let the embedding dimension of the student model’s deepest layer output be $d_s$, where $d_t >> d_s$. The codebook consists of  N vectors, each of M dimensions, where $M = d_s$. Therefore, during the teacher model’s pretraining phase, we employ a dimension alignment module to map the teacher model’s encoder output to $d_s$ dimensions and subsequently map the discrete vectors returned by the codebook back to $d_t$ dimensions. In the knowledge distillation stage, the student model’s deepest layer output learns to align as closely as possible with the discrete outputs from the codebook.

This approach offers two additional benefits. First, reducing the dimensionality of the codebook enhances its utilization and alleviates the issue of codebook collapse, which can occur when the codebook is overly complex\cite{peng2022beit}.Second, the codebook effectively involves a k-means clustering process; clustering results in high dimensions can be confounded by irrelevant and redundant dimensions \cite{parsons2004subspace}. Lowering the dimensionality of the codebook aids in generating a higher-quality codebook.

Following the methodology of VQ-Wav2Vec \cite{baevski2019vq}, we employ Gumbel-Softmax to mitigate the issue of codebook collapse. To achieve dimensional alignment, we evaluate the performance of linear projection, 1x1 convolution, and channel selection. Due to comparable performance, we selected 1x1 convolution for this application.  Thus, the output of the codebook can be expressed as :
\begin{equation}
z_{i}^{'}   = \underset{j}{Gumbel Softmax} \parallel DA(z_i)-v_j\parallel_{2 } * DA(V)
\end{equation}
Here, "DA" denotes dimension alignment.
We train the codebook using the same loss function as VQ-VAE, and apply an L1 loss to enforce consistency between the student model’s encoder output and the discrete outputs of the codebook. Consequently, the codebook loss can be formulated as: 
\begin{equation}
\mathcal{L}_{\text{cb}}^T = \left\| \text{sg}[DA] -z_{i}^{'} \right\|_2^2 + \lambda \left\| DA(z_i) - \text{sg}[z_{i}^{'}] \right\|_2^2
\end{equation}

\subsection{The Knowledge Distillation Stage}
In the knowledge distillation stage, we integrate semi-supervised learning with intermediate layer information distillation to optimize training. The student model is designed as a lightweight CNN, aiming to minimize parameter count and reduce computational complexity.

The overall loss function incorporates multiple objectives to enhance model performance, expressed as follows:
\begin{align}
\footnotesize
L_{stu} &= L_{seg}^S + \beta L_{semi}^S + \gamma L_{dis}^S \\
&= DICE(Y_{stu}, Y_{gt}) + \beta DICE(Y_{stu}, Y_{pse}) \\
&\quad + \gamma L1(Z_{stu}, Z_{tea}^{'})
\end{align}

The first component represents the segmentation loss derived from a limited number of labeled examples, $Y_{gt}$. The second component captures the segmentation loss from the pseudo-labels $Y_{pse}$, generated by the teacher model for unlabeled data. The final component is the distillation loss, which aligns the intermediate layer outputs  $Z_{stu}$  of the student model with the discretized outputs $Z_{tea}^{'}$ from the teacher model's codebook.

By leveraging semi-supervised learning and distilling information from intermediate layers, we aim to enhance the generalization and information extraction capabilities of the student model.

\begin{table*}[t]
\label{table:Comprision}
\centering
\caption{Quantitative Comparison of Different Methods on Cerebrovascular Segmentation Using Five-Fold Cross Validation}
\begin{tabular}{ccccccc}
\hline
\rowcolor[HTML]{EFEFEF} 
\textbf{Method}   & \textbf{DSC($\uparrow$)} & \textbf{Accuracy($\uparrow$)} & \textbf{HD95($\downarrow$)} & \textbf{Jaccard($\uparrow$)} & \textbf{GWD($\downarrow$)} & \textbf{Cl\_Dice($\uparrow$)} \\ \hline

\multicolumn{7}{c}{\textbf{Full Supervision Baseline Method}} \\ \hdashline
Light UNet & 0.823{\footnotesize\textcolor{gray}{$\pm$0.046}} & 0.970{\footnotesize\textcolor{gray}{$\pm$0.007}} & 2.711{\footnotesize\textcolor{gray}{$\pm$1.827}} & 0.710{\footnotesize\textcolor{gray}{$\pm$0.060}} & 0.261{\footnotesize\textcolor{gray}{$\pm$0.034}} & 0.873{\footnotesize\textcolor{gray}{$\pm$0.041}} \\ \hline

\multicolumn{7}{c}{\textbf{Knowledge Distillation Based Methods}} \\ \hdashline
Vanilla KD & 0.834{\footnotesize\textcolor{gray}{$\pm$0.025}} & 0.971{\footnotesize\textcolor{gray}{$\pm$0.006}} & 2.409{\footnotesize\textcolor{gray}{$\pm$1.315}} & 0.723{\footnotesize\textcolor{gray}{$\pm$0.033}} & 0.257{\footnotesize\textcolor{gray}{$\pm$0.029}} & 0.882{\footnotesize\textcolor{gray}{$\pm$0.029}} \\ 
Fitnets & 0.842{\footnotesize\textcolor{gray}{$\pm$0.030}} & 0.973{\footnotesize\textcolor{gray}{$\pm$0.006}} & 2.244{\footnotesize\textcolor{gray}{$\pm$1.258}} & 0.739{\footnotesize\textcolor{gray}{$\pm$0.042}} & 0.243{\footnotesize\textcolor{gray}{$\pm$0.026}} & 0.889{\footnotesize\textcolor{gray}{$\pm$0.034}} \\ 
VID & 0.846{\footnotesize\textcolor{gray}{$\pm$0.019}} & 0.973{\footnotesize\textcolor{gray}{$\pm$0.006}} &  2.184{\footnotesize\textcolor{gray}{$\pm$0.817}} &  0.741{\footnotesize\textcolor{gray}{$\pm$0.025}} &  0.256{\footnotesize\textcolor{gray}{$\pm$0.025}} & 0.885{\footnotesize\textcolor{gray}{$\pm$0.022}} \\ 
AB & 0.846{\footnotesize\textcolor{gray}{$\pm$0.032}} & 0.974{\footnotesize\textcolor{gray}{$\pm$0.005}} & 2.134{\footnotesize\textcolor{gray}{$\pm$1.181}} & 0.744{\footnotesize\textcolor{gray}{$\pm$0.045}} & 0.240{\footnotesize\textcolor{gray}{$\pm$0.028}} & 0.889{\footnotesize\textcolor{gray}{$\pm$0.034}} \\ \hline

\multicolumn{7}{c}{\textbf{Semi-supervision Based Methods}} \\ \hdashline
MT & 0.839{\footnotesize\textcolor{gray}{$\pm$0.037}} & 0.973{\footnotesize\textcolor{gray}{$\pm$0.005}} & 2.208{\footnotesize\textcolor{gray}{$\pm$1.178}} & 0.732{\footnotesize\textcolor{gray}{$\pm$0.050}} & 0.242{\footnotesize\textcolor{gray}{$\pm$0.026}} & 0.892{\footnotesize\textcolor{gray}{$\pm$0.033}} \\ 
UAMT & 0.847{\footnotesize\textcolor{gray}{$\pm$0.029}} & 0.974{\footnotesize\textcolor{gray}{$\pm$0.005}} & 2.337{\footnotesize\textcolor{gray}{$\pm$1.473}} & 0.743{\footnotesize\textcolor{gray}{$\pm$0.039}} & 0.241{\footnotesize\textcolor{gray}{$\pm$0.022}} & 0.892{\footnotesize\textcolor{gray}{$\pm$0.031}} \\ 
ME & 0.847{\footnotesize\textcolor{gray}{$\pm$0.024}} & 0.974{\footnotesize\textcolor{gray}{$\pm$0.006}} & 2.457{\footnotesize\textcolor{gray}{$\pm$1.546}} & 0.749{\footnotesize\textcolor{gray}{$\pm$0.040}} & 0.235{\footnotesize\textcolor{gray}{$\pm$0.023}} & 0.899{\footnotesize\textcolor{gray}{$\pm$0.029}} \\ \hline

\textbf{VpbSD(Ours)}     & 0.852{\footnotesize\textcolor{gray}{$\pm$0.026}}  & 0.974{\footnotesize\textcolor{gray}{$\pm$0.006}} & 1.974{\footnotesize\textcolor{gray}{$\pm$0.804}} & 0.751{\footnotesize\textcolor{gray}{$\pm$0.035}} & 0.231{\footnotesize\textcolor{gray}{$\pm$0.022}} & 0.900{\footnotesize\textcolor{gray}{$\pm$0.025}} \\ \hline
\end{tabular}
\end{table*}

\iffalse
 SwinUNetR(Teacher) & 0.875±0.037 & 0.978±0.008 & 1.942±1.288 & 0.789±0.053 & 0.203±0.032 & 0.913±0.033 &             63.98M \\
        UnetR & 0.852±0.034 & 0.973±0.007 & 1.607±0.583 & 0.766±0.053 & 0.221±0.034 & 0.910±0.035 &            98.186M \\ \hline
\fi
%
\begin{figure*}[t] % [t]表示图片顶部对齐，可以根据需要调整
    \centering
    \includegraphics[width=\textwidth]{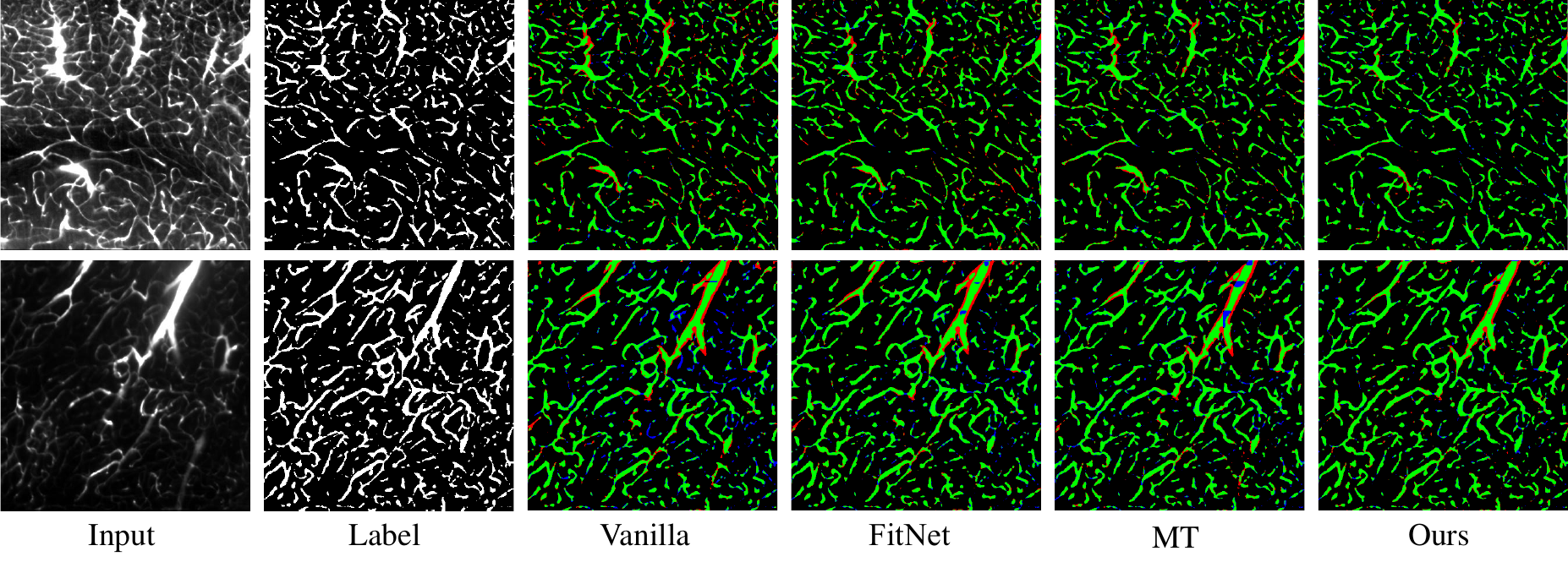} % 将图片宽度设置为文本宽度
    \caption{Visual comparative segmentation results between the proposed methods and existing approaches. Each example presents the input image alongside the ground truth (label) and the corresponding segmentation outcomes. Green regions denote accurate segmentations, blue regions indicate missed detections, and red regions signify over-segmentations. Two representative examples are provided for thorough evaluation.}
    \label{fig:Seg_Result_Illustration}
 \vspace{-1.5em}
\end{figure*}
\section{Experiments}
\label{sec:exp}
\subsection{Implementation Details}
\subsubsection{Datasets}

In our experiments, we utilized the public VesSep2020 dataset, introduced in 2020 \cite{todorov2020machine}. This dataset, derived from light-sheet microscopy, comprises a substantial collection of unlabeled microscopic images of mouse brains, complemented by a smaller subset of manually annotated data. The resolution of VesSep2020 is 1.625 \(\times\) 1.625 \(\times\) 3.0 \(\mu m^3\) featuring 11 manually annotated samples, each with dimensions of 500×500×50. The dataset encompasses 3D microscopic images of cerebrovascular structures across four different mouse strains, totaling 12 subjects. The images are presented in a 2D atlas format, utilizing two distinct staining agents. Given the limited availability of 11 manually labeled samples, we conducted a 5-fold cross-validation.
\iffalse
To minimize the influence of individual data characteristics, we divided the data outside the training set into non-overlapping 50x50x50 blocks, which were then randomly shuffled. These blocks were split in a 6:4 ratio to form the test and validation sets, each containing approximately 100 blocks. This approach ensured that both the test and validation sets were representative and independent of each other. \fi

For the unlabeled 2D microscopic images, we employed Terastitcher \cite{bria2012terastitcher} to stitch the cerebrovascular microscopic slices of each mouse into 3D volumes. These volumes were subsequently divided into equal-sized 3D patches and stored on disk, reducing memory and GPU requirements during training. Prior to model input, we performed histogram equalization to analyze the distribution characteristics. This involved retaining the values below the 95th percentile and truncating any values above this threshold. The data was normalized using min-max scaling to mitigate the impact of extreme values on model performance. Although the original authors provided vascular staining results from the same mouse brain using two different agents, we selected the dataset with only the wheat germ agglutinin agent for our study due to computational limitations, thereby demonstrating the effectiveness of our method.

\subsubsection{Models}
For the teacher model, we modified the Swin UNETR architecture \cite{tang2022self}, leveraging it as the baseline to align with our experimental design. This model exemplifies a state-of-the-art 3D segmentation approach, integrating a Transformer-based encoder \cite{vaswani2017attention} with a CNN-based decoder.

For the student model, we employed a lightweight version of UNet, comprising four layers in both the encoder and decoder. Each encoder layer consists of two convolutional kernels, followed by Group Normalization and Leaky ReLU activation. The final operation in each encoder layer is downsampling, which reduces the spatial dimensions of the feature maps. The decoder mirrors this structure, utilizing two convolutional kernels per layer, each followed by Group Normalization and Leaky ReLU. Additionally, residual connections are incorporated between corresponding layers of the encoder and decoder to facilitate improved feature propagation and reconstruction.

All models were constructed within the PyTorch framework.
\begin{table*}[htb]
\label{table:FLOPS}
\centering
\setlength{\tabcolsep}{4pt} % Adjust column padding if needed
\caption{Comparison of Model Parameters, FLOPs, Inference Time, and Performance Metrics}
\begin{tabular}{cccccccc}
\hline
\rowcolor[HTML]{EFEFEF} 

\textbf{Model} & \textbf{DSC($\uparrow$)} & \textbf{HD95($\downarrow$)}& \textbf{GWD($\downarrow$)}  & \textbf{Cl\_Dice($\uparrow$)}  & \textbf{GFLOPs} & \textbf{Parameters (M)} & \textbf{Time (ms)} \\ \hline
\rowcolor[HTML]{E7E6E6} SwinUNetR & 0.875 & 1.942 & 0.203 & 0.913 & 98.63 & 63.98 & 22.63 \\
UNetR & 0.852 & 1.607 & 0.221 & 0.910 & 273.84 & 98.19 & 24.26 \\

\rowcolor[HTML]{E7E6E6} FitNet & 0.842 & 2.244 & 0.243 & 0.889 & 0.11 & 0.16 & 2.70 \\
VID & 0.846 &  2.184 &  0.256 & 0.885 & 0.12 & 0.59 & 2.83 \\ \hline
\rowcolor[HTML]{E7E6E6} \textbf{VpbSD(Ours)} & 0.852 & 1.974 & 0.231 & 0.900 & 0.11 & 0.12 & 2.49 \\ \hline
\end{tabular}
\end{table*}

\subsubsection{Experiments Settings}
All experiments were conducted on A100 GPUs with 48GB of memory. We utilized the Adam optimizer with an exponentially decaying learning rate. During training, the 3D blocks were divided into patches of size 64x64x64 and processed in batches of 2. The teacher model underwent pretraining for 80 epochs, followed by fine-tuning for 30 epochs. All knowledge distillation experiments were conducted over 300 epochs. The initial learning rates for these three stages were set to 1e-3, 1e-5, and 1e-2, respectively. Hyperparameters were configured with $\alpha$ set to 0.1 and $\gamma$ to 1, while $\beta$ was updated using a commonly employed warming-up function \cite{laine2016temporal}. The project code is available at https://github.com/lwannnn.
\subsubsection{Evaluation metrics}
To comprehensively assess the model's performance, we employed multiple metrics categorized into overlap-based and distance-based measures. The overlap-based metrics included the Dice Similarity Coefficient (DSC), accuracy, Jaccard index, and cldice \cite{shit2021cldice}, which evaluate the similarity between predicted and ground truth regions. Distance-based metrics, such as Hausdorff Distance at 95\% (HD95) and Skeleton Matching Distance (SMD) \cite{lin2024vascular}, quantify skeletal displacement between vessels. The following outlines the calculation methods for each evaluation metric used in this study:

DSC: Measures the overlap between predicted and actual regions, emphasizing segmentation accuracy.
\begin{equation}
DICE(X,Y) = \frac{2\left | X\cap Y \right |  }{\left | X \right | +\left | Y \right | } 
\end{equation}
where $X$ and $Y$ represent the predicted and ground truth sets, respectively. 

Accuracy: Reflects the proportion of correctly classified elements across the entire prediction.
\begin{equation}
Acc = \frac{TP + TN}{TP + TN + FP + FN}
\end{equation}
where $TP,TN, FP,$ and $FN $ represent true positives, true negatives, false positives, and false negatives.

Jaccard: Provides a ratio of the intersection over the union of the predicted and actual regions, measuring set similarity.
\begin{equation}
Jaccard = \frac{|X \cap Y|}{|X \cup Y|}
\end{equation}
cldice: A refinement of the Dice coefficient, focusing on skeleton similarity, particularly for tubular segmentation tasks.

HD95: Captures the maximum deviation between the predicted and actual boundaries, ignoring extreme outliers.
\begin{equation}
HD95 = \max \left\{ \min_{x \in X} d(x, Y), \min_{y \in Y} d(y, X) \right\}_{95th}
\end{equation}
SMD: SMD is an evaluation metric based on Sinkhorn Distance\cite{cuturi2013sinkhorn}, which quantifies the minimum cost required to transform one distribution into another, effectively capturing spatial deviations. Leveraging this, SMD is designed to assess the degree of skeletal displacement between inferred vessels and ground truth vessels.

\begin{table*}[t]
\label{table:Ablation}
\centering
\setlength{\tabcolsep}{4pt}
\caption{Performance Comparison in Ablation Study for Key Component Analysis}
\begin{tabular}{cccccccc}
\hline
\rowcolor[HTML]{EFEFEF} 
\textbf{Method}   & \textbf{DSC($\uparrow$)} & \textbf{Accuracy($\uparrow$)} & \textbf{HD95($\downarrow$)} & \textbf{Jaccard($\uparrow$)} & \textbf{GWD($\downarrow$)} & \textbf{Cl\_Dice($\uparrow$)}  \\ \hline
SwinUNetR(Teacher) & 0.875{\footnotesize\textcolor{lightgray}{$\pm$0.037}} & 0.978{\footnotesize\textcolor{lightgray}{$\pm$0.008}} & 1.942{\footnotesize\textcolor{lightgray}{$\pm$1.288}} & 0.789{\footnotesize\textcolor{lightgray}{$\pm$0.053}} & 0.203{\footnotesize\textcolor{lightgray}{$\pm$0.032}} & 0.913{\footnotesize\textcolor{lightgray}{$\pm$0.033}}  \\ 
Light UNet(Baseline) & 0.823{\footnotesize\textcolor{lightgray}{$\pm$0.046}} & 0.970{\footnotesize\textcolor{lightgray}{$\pm$0.007}} & 2.711{\footnotesize\textcolor{lightgray}{$\pm$1.827}} & 0.710{\footnotesize\textcolor{lightgray}{$\pm$0.060}} & 0.261{\footnotesize\textcolor{lightgray}{$\pm$0.034}} & 0.873{\footnotesize\textcolor{lightgray}{$\pm$0.041}} \\ \hline
VpbSD w/o $L_{dis}^S$ & 0.842{\footnotesize\textcolor{lightgray}{$\pm$0.033}} & 0.973{\footnotesize\textcolor{lightgray}{$\pm$0.007}} & 2.126{\footnotesize\textcolor{lightgray}{$\pm$1.166}} & 0.738{\footnotesize\textcolor{lightgray}{$\pm$0.045}} & 0.243{\footnotesize\textcolor{lightgray}{$\pm$0.028}} & 0.889{\footnotesize\textcolor{lightgray}{$\pm$0.035}}  \\ 
VpbSD w/o $L_{semi}^S$ & 0.846{\footnotesize\textcolor{lightgray}{$\pm$0.024}} & 0.973{\footnotesize\textcolor{lightgray}{$\pm$0.007}} & 2.060{\footnotesize\textcolor{lightgray}{$\pm$0.749}} & 0.743{\footnotesize\textcolor{lightgray}{$\pm$0.031}} & 0.238{\footnotesize\textcolor{lightgray}{$\pm$0.018}} & 0.895{\footnotesize\textcolor{lightgray}{$\pm$0.022}}  \\ \hline
Baseline +$L_{mid}$ & 0.842{\footnotesize\textcolor{gray}{$\pm$0.030}} & 0.973{\footnotesize\textcolor{gray}{$\pm$0.006}} & 2.244{\footnotesize\textcolor{gray}{$\pm$1.258}} & 0.739{\footnotesize\textcolor{gray}{$\pm$0.042}} & 0.243{\footnotesize\textcolor{gray}{$\pm$0.026}} & 0.889{\footnotesize\textcolor{gray}{$\pm$0.034}} \\ 
Baseline +$L_{semi}^S$ +$L_{mid}$ & 0.844{\footnotesize\textcolor{gray}{$\pm$0.030}} & 0.973{\footnotesize\textcolor{gray}{$\pm$0.006}} & 2.244{\footnotesize\textcolor{gray}{$\pm$1.258}} & 0.741{\footnotesize\textcolor{gray}{$\pm$0.042}} & 0.243{\footnotesize\textcolor{gray}{$\pm$0.026}} & 0.891{\footnotesize\textcolor{gray}{$\pm$0.032}} \\
 \hline\hline

VpbSD & 0.852{\footnotesize\textcolor{lightgray}{$\pm$0.026}} & 0.974{\footnotesize\textcolor{lightgray}{$\pm$0.006}} & 1.974{\footnotesize\textcolor{lightgray}{$\pm$0.804}} & 0.751{\footnotesize\textcolor{lightgray}{$\pm$0.035}} & 0.231{\footnotesize\textcolor{lightgray}{$\pm$0.022}} & 0.900{\footnotesize\textcolor{lightgray}{$\pm$0.025}}  \\ \hline
\end{tabular}
\end{table*}

\subsection{Comparative Experiments}

\subsubsection{Comparative Evaluation of Strategy-Driven Performance Enhancement}
We compared our VpbSD method against several state-of-the-art approaches, specifically selecting well-established knowledge distillation techniques, including vanilla KD \cite{hinton2015distilling}, FitNets \cite{romero2014fitnets}, VID \cite{ahn2019variational}, and AB \cite{heo2019knowledge}. Additionally, we included prominent semi-supervised learning methods such as Mean Teacher (MT) \cite{tarvainen2017mean}, UAMT \cite{yu2019uncertainty}, and Advent \cite{vu2019advent}.

In terms of experimental design, we adhered to the original configurations wherever specific setups were provided, ensuring consistency in parameters such as learning rate, number of epochs, and batch size for all others. Notably, all knowledge distillation algorithms utilized the same architecture for both teacher and student models, while semi-supervised methods relied on a uniform student model architecture. Our objective was to compare these strategies by evaluating performance enhancements of the baseline model without altering its structure.

As detailed in Table \ref{table:Comprision}, VpbSD demonstrates significant improvements in segmentation performance using the same baseline model. Specifically, VpbSD outperforms all listed state-of-the-art methods, effectively enhancing the baseline model across various metrics. The Cl\_Dice score increased from 0.873 to 0.900, while DSC rose to 0.852, and Jaccard improved to 0.751. Notably, HD95 decreased to 1.974, indicating enhanced boundary precision. Furthermore, VpbSD achieved a lower GWD of 0.231, reflecting its capability to preserve vascular morphological consistency. These results highlight that our model not only surpasses both semi-supervised and knowledge distillation approaches but also significantly boosts the segmentation accuracy of the baseline model under consistent experimental conditions.  In  Fig~\ref{fig:Seg_Result_Illustration}, we present segmentation results from two 2D perspectives to facilitate an intuitive comparison of performance across different methods. The visual results indicate that VpbSD effectively reduces excessive and missed segmentations compared to other models.

\subsubsection{Comparative Study on Balancing Performance and Computational Efficiency in Model Deployment}

In the task of 3D brain vascular segmentation, the high resolution associated with large data volumes amplifies even minor differences in model inference speed. This underscores the critical need to balance model accuracy with practical deployment, particularly in scenarios demanding rapid inference. To evaluate the trade-offs between computational efficiency and performance across various methods, we compared their computational complexity in terms of GFLOPs, parameter count, and inference time. Additionally, we presented metrics such as DSC, HD95, GWD, and cl\_dice in the accompanying table to assess the relative performance and resource utilization of the models. Inference time was measured on an A100 GPU with 48 GB of memory, where the model processed 100 blocks of dimensions 64 × 64 × 64, with the table listing the average time required to process a single block.

Our comparison included two Transformer-based methods: SwinUNetR\cite{tang2022self} and UNetR\cite{hatamizadeh2022unetr}, with SwinUNetR utilized as our teacher model. We also incorporated FitNet\cite{romero2014fitnets} and VID\cite{ahn2019variational}, which introduce new modules into the baseline models.

As illustrated in Table \ref{table:FLOPS}, the segmentation performance of VpbSD is comparable to that of UNetR; however, UNetR requires a parameter count of 98.63M, while VpbSD utilizes only 0.12M parameters. The strategies employed by FitNet and VID result in a slight increase in computational cost and resource consumption; nonetheless, our approach demonstrates a more pronounced improvement over the baseline models. Although SwinUNetR, as our teacher model, exhibits strong segmentation performance, it has a GFLOPs value of 98.63, and its inference time per patch is nearly 11 times greater than that of our proposed method. This significant increase in inference time presents a notable drawback when segmenting a substantial number of microscopic brain vessels. Consequently, VpbSD emerges as a robust solution for microscopic cerebral vascular segmentation, rendering it particularly well-suited for real-time processing in high-resolution applications.

\subsection{Ablation Study}
In this section, we also conducted an ablation study to analyze the effectiveness of each module. We compared the model's performance after removing different strategies and examined the impact of various dimension alignment strategies on the final segmentation results.

\subsubsection{Analysis of Strategy Effectiveness}
In VpbSD, we integrate the principles of semi-supervised learning along with a vessel pattern-based knowledge distillation strategy. This approach aims to enhance the generalization ability of the student model while alleviating the dependency on labeled data. To demonstrate the effectiveness of these strategies, we compared the combinations of these approaches as shown in Table \ref{table:Ablation}. First, we presented the segmentation results obtained by removing $L_{dis}^S$ (VpbSD w/o $L_{dis}^S$) or $L_{semi}^S$(VpbSD w/o $L_{semi}^S$) from VpbSD. This analysis highlighted the contributions of each strategy to the model's performance enhancement. We observed that the semi-supervised strategy and the use of intermediate feature distillation alone improved performance by 1.9\% and 2.5\%, respectively, compared to the baseline in terms of DSC.  The combination of these strategies yielded the best results, leading to improvements of +2.9\%, +0.4\%, -0.737, +4.1\%, -0.03, and +2.7\% for DSC, accuracy, HD95, Jaccard, GWD, and Cl\_Dice, respectively. These experimental results indicated that each strategy significantly enhanced the model's segmentation performance, demonstrating the overall effectiveness of our approach.

Furthermore, we aimed to show that the improvement attributed to  $L_{dis}^S$ resulted specifically from vessel pattern-based distillation rather than solely from enhancements provided by intermediate layer knowledge distillation. We presented training results obtained by adding $L_{mid}$  to the baseline model and compared this group with the results of "VpbSD w/o $L_{semi}^S$". Here, $L_{mid} = L1(DA(Z_{stu}), Z_{tea})$. We noted that replacing vessel pattern-based knowledge distillation with standard intermediate layer knowledge distillation led to a decrease of 0.004 in the DSC score, a reduction of 0.006 in Cl\_Dice, and an increase of 0.005 in GWD. This finding suggested that vessel pattern-based knowledge distillation better preserved the vascular structure information learned by the student model, enabling it to acquire a more diverse set of vascular morphological features.
Additionally, we presented training results obtained by incorporating both $L_{mid}$ and $L_{semi}^S$ into the baseline model, comparing this group with VpbSD. We observed that VpbSD significantly improved the DSC score 0.008 compared to "Baseline +$L_{mid}$ + $L_{semi}^S$ ". Additionally, the Cl\_dice metric increased from 0.891 to 0.900, further  demonstrating that vessel pattern-based knowledge distillation effectively enhanced the segmentation performance of VpbSD.

\iffalse
\subsubsection{Analysis of Vascular Pattern Capacity}
We believe that codebook-based knowledge distillation can provide the student model with richer and more structured knowledge. Therefore, the size of the codebook may be an important parameter. If the size is too small, it may not encompass diverse vascular structure information; if it is too large, it may lead to more fragmented clustering results. To investigate this, we compared the impact of different codebook sizes on the model's segmentation results. We set the codebook sizes to 256, 512, 1024, and 2048, with the final results shown in Figure x. Additionally, we used t-SNE to examine the distribution and utilization of the codebook, as illustrated in Figure x.
\fi

\section{Conclusion}
\label{sec:clu}
In this paper, we propose a semi-supervised knowledge distillation model integrated with vector quantization to  tackle the challenge of microscopic brain vessel segmentation. Our strategies effectively leveraged feature learning from difficult-to-annotate microscopic images, resulting in a student model capable of fast and efficient segmentation. This approach alleviated the computational burden typically associated with microscopic image analysis. We conducted extensive experiments, providing both quantitative and visual comparisons that demonstrate the effectiveness and superiority of our method.

\bibliographystyle{IEEEbib}
\bibliography{strings,refs}

\end{document}